\pdfoutput=1

\documentclass[11pt]{article}

\usepackage{naacl2021}

\usepackage{times}
\usepackage{latexsym}
\usepackage[T1]{fontenc}
\usepackage[utf8]{inputenc}

\usepackage{microtype}

\usepackage{amsmath}
\usepackage{enumitem}
\usepackage[leftmargin=5mm,rightmargin=5mm]{quoting}
\usepackage{url}
\usepackage{booktabs}
\usepackage{amssymb}
\usepackage{graphicx}
\usepackage{multirow}
\usepackage{scalerel}
\usepackage{boxedminipage}
\usepackage{color, colortbl}
\definecolor{LightCyan}{rgb}{0.88,1,1}
\definecolor{LightRed}{rgb}{1.0, 0.91, 0.91}
\definecolor{LightGray}{rgb}{0.88,0.88,0.88}
\definecolor{VeryLightGray}{rgb}{0.93,0.93,0.93}

\RequirePackage{fix-cm} 

\title{Self-Supervised Test-Time Learning for Reading Comprehension}

\author{
Pratyay Banerjee \quad Tejas Gokhale \quad Chitta Baral  \\
Arizona State University \\
\texttt{pbanerj6, tgokhale, chitta}@asu.edu               \\
}

\begin{document}
\maketitle
\begin{abstract}
    Recent work on unsupervised question answering has shown that models can be trained with procedurally generated question-answer pairs and can achieve performance competitive with supervised methods. 
    In this work, we consider the task of unsupervised reading comprehension and present a method that performs ``test-time learning'' (TTL) on a given context (text passage), 
    without requiring training on large-scale human-authored datasets 
    containing \textit{context-question-answer} triplets. 
    This method operates directly on a single test context, uses self-supervision to train models on synthetically generated question-answer pairs, and then infers answers to unseen human-authored questions for this context. Our method achieves accuracies competitive with fully supervised methods and significantly outperforms current unsupervised methods.
    TTL methods with a smaller model are also competitive with the current state-of-the-art in unsupervised reading comprehension.
\end{abstract}

\section{Introduction}

Reading comprehension is the task in which systems attempt to answer questions about a passage of text.
Answers are typically found in the passage as text-spans or can be inferred through various forms of reasoning~\cite{rajpurkar-etal-2016-squad}.
The answer to the following question:

\begin{quoting}
    \noindent{\it``Who is the President of the United States?''}
\end{quoting}
\noindent depends on the timeframe and context of the passage provided, and will be different for news articles written in 2001 vs. 2021.
If the context is the script of the TV series ``The West Wing'', the answer is ``Jed Bartlet'', and even in this fictional setting, it will later change to ``Matt Santos''.

Knowledge sources such as Wikipedia get updated when new events occur (such as the outcome of elections), or new facts about the world are revealed (such as scientific discoveries), with contributors adding new information and removing information that is no longer valid~\cite{almeida2007evolution}.
With such context-dependent answers and continual changes in knowledge, it is hard to justify training models over fixed corpora for tasks such as question answering (QA).
We would like models to answer questions based on the given context and not to learn biases from datasets or historical news articles.

Moreover, supervised learning has been shown to perform poorly in QA tasks with adversarial examples~\cite{jia-liang-2017-adversarial}, domain shift~\cite{jia-liang-2017-adversarial,yogatama2019learning,kamath-etal-2020-selective}, and biased or imbalanced data~\cite{agrawal2018don,mccoy-etal-2019-right}.
For example, QA systems trained on Wikipedia fail to generalize to newer domains such as Natural Questions~\cite{rennie-etal-2020-unsupervised} or biomedical data~\cite{wiese-etal-2017-neural}, and suffer a significant drop in accuracy.
Even small semantics-preserving changes to input sentences, such as the substitution of words by synonyms, have been shown to degrade performance in NLP tasks~\cite{alzantot-etal-2018-generating,jia-etal-2019-certified}.
Continual changes in text corpora are inevitable, thus calling for the development of robust methods that can reliably perform inference without being subject to biases.

Supervised Question Answering faces challenges such as the need for large-scale (usually human-authored) training corpora to train models.
Such corpora typically require significant post-processing and filtering to remove annotation artifacts~\cite{sakaguchi2020winogrande}.
To address these challenges, some recent methods~\cite{lewis-etal-2019-unsupervised,li-etal-2020-harvesting} approach question answering as an unsupervised learning task.
A significant advantage of this approach is that it can be extended to domains and languages for which collecting a large-sized human-authored training corpus is challenging. 
Methods for unsupervised QA procedurally generate a large corpus of {\it (context, question, answer)} triples, and train large neural language models, such as BERT~\cite{devlin-etal-2019-bert}.

In this work, we focus on unsupervised reading comprehension (RC) under evolving contexts and present the ``Test-Time Learning" paradigm for this task.
RC -- the task of answering questions about a passage of text,
acts as the perfect setting for robust question-answering systems that do not overfit to training data.
While large-scale language models trained on large datasets may contain global information, the answer needs to be extracted from the given context.
Thus, our work seeks to learn unsupervised reading comprehension without access to human-authored training data but instead operates independently on each test context.
This makes our method `distribution-blind' where each new context is assumed to be a novel distribution.
The test-time learning (TTL) framework enables smaller models to achieve improved performance with small procedurally generated question-answer pairs,  
and is summarized below:

\noindent\begin{boxedminipage}{\linewidth}
\begin{itemize}[leftmargin=*,nosep,noitemsep]
    \item a single context (text passage) $c_i$ is given, from which we procedurally generate QA pairs;
    \item these QA pairs are used to train models to answer questions about $c_i$;
    \item the inference is performed on previously unseen questions for $c_i$.
\end{itemize}
\end{boxedminipage}

This framework has a simple assumption that every context comes from a distinct distribution.
Hence, parameters learned for the previous context might not be useful to generalize to other contexts.
This assumption holds where the contexts evolve over time, and rote memorization of answers might lead to wrong predictions.
As such, the above process is repeated for each new context $c_i$.

For question-answer generation, we use simple methods such as cloze-translation~\cite{lewis-etal-2019-unsupervised}, template-based question-answer generation~\cite{fabbri-etal-2020-template} and question-answer semantic role labeling (QA-SRL)~\cite{he-etal-2015-question}. 
We use two neural transformer-based language models, BERT-Large~\cite{devlin-etal-2019-bert} and DistilBert~\cite{sanh2019distilbert}, to study the efficacy of our framework with large and small transformer models.
We evaluate our method on two reading comprehension datasets, SQuAD~\cite{rajpurkar-etal-2016-squad} and NewsQA~\cite{trischler-etal-2017-newsqa}. 
We investigate test-time training under multiple learning settings:
(1) single-context learning -- the ``standard'' setting, 
(2) $K$-neighbor learning -- by retrieving top-$K$ multiple related contexts for each test context, 
(3) curriculum learning -- progressively learning on question-types of increasing order of complexity, 
(4) online learning -- sequentially finetuning models on each incoming test sample.

\noindent{}

\noindent Our experimental findings are summarized below:
\begin{itemize}[nosep,noitemsep]
    \item Test-time learning methods are effective for the task of reading comprehension and surpass current state-of-the-art on two benchmarks: SQuAD and NewsQA.
    \item Online TTL trained over K-neighboring contexts of the test context is the best version with EM/F1 gains of $7.3\% / 7.8\%$ on SQuAD 1.1 and $5.3\% / 6.9\%$ on NewsQA.
    \item DistilBERT -- which has less than $\frac{1}{5}^{th}$ of the number of model parameters of BERT-Large is competitive with current SOTA methods that use BERT-Large.
\end{itemize}



\begin{figure*}[t]
    \centering
    \includegraphics[width=\linewidth]{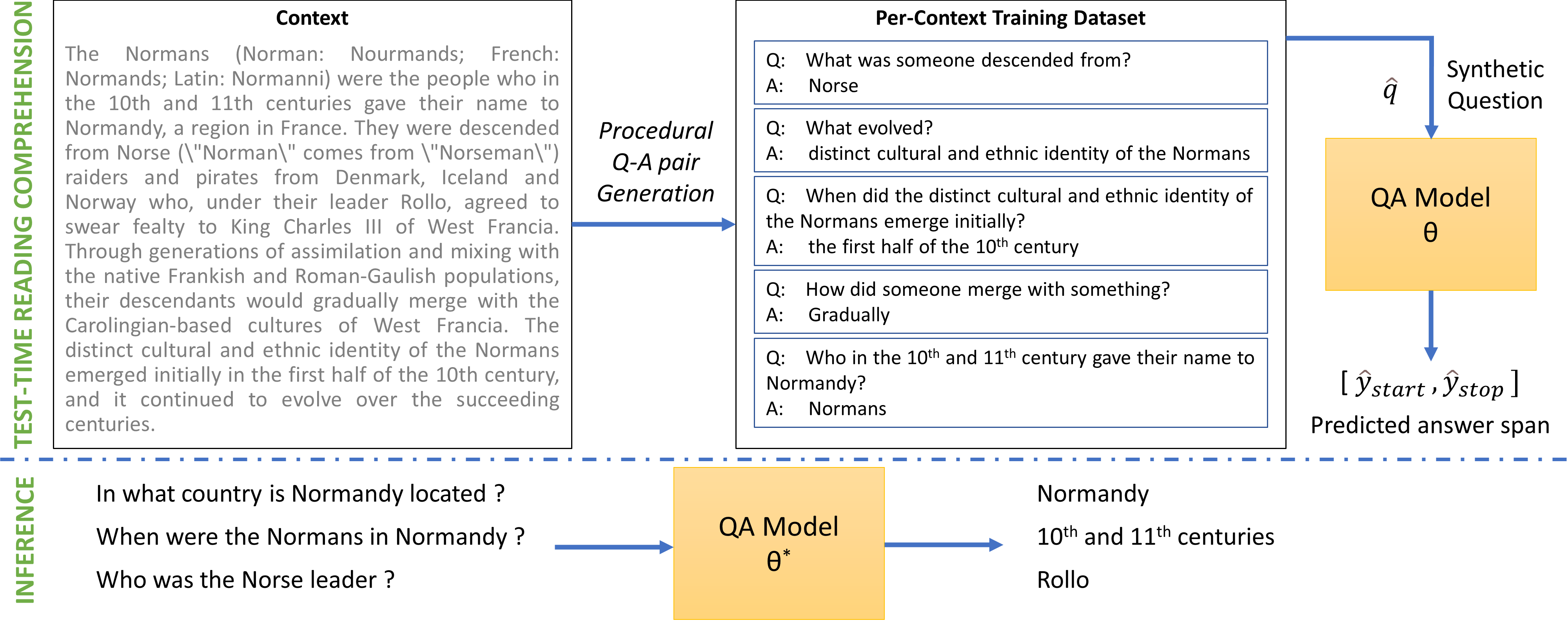}
    \caption{
    Overview of our self-supervised test-time learning framework for reading comprehension. 
    Our method does not require a human-authored training dataset but operates directly on each single test context and synthetically generates question-answer pairs over which model parameters $\theta$ are optimized.
    The inference is performed with trained parameters $\theta^*$ on unseen human authored questions.}
    \label{fig:overview}
\end{figure*}

\section{Test-Time Reading Comprehension}
Consider a reading comprehesion test dataset $\mathcal{D}^{test}{=}\{ (c_i, q_i, a_i)\}_{i=1}^{n}$ with context text passages $c_i$, human-authored questions $q_i$ and true answers $a_i$.
The QA model $g(\cdot)$ is parameterized by $\theta = (\theta_f, \theta_h)$ where $\theta_f$ are parameters for the feature extractor, and $\theta_h$ for the answering head.
The answer is predicted as a  text-span, given by the start and stop positions $[y_{start}, y_{stop}]$.
Contemporary unsupervised RC models \cite{lewis-2019-compositional,li-etal-2020-harvesting} are trained on a large dataset $\mathcal{\hat{D}}^{train}{=}\{ (c_i, \hat{q}_i, \hat{a}_i)\}_{i=1}^{n}$, where the QA pairs are synthetically generated from the context.

In our setting, we do not use such large training datasets, but instead directly operate on individual test contexts $c_i \in \mathcal{D}^{test}$.
Given $c_i$, M synthetic question-answer pairs $\{(\hat{q}_i^j, \hat{a}_i^j)\}_{j=1}^M$ are procedurally generated as described in Section~\ref{selfsup_qa_gen}.
The QA model 
parameters $\theta$ are 
trained over the synthetic data to predict the span of the answer $[\hat{y}_{start},\hat{y}_{stop}]$
by optimizing the loss $\ell_{ans}$:
\begin{gather}
    \small
    \underset{\theta}{\text{minimize}}~~\sum_{j=1}^{M} \ell_{ans}(c_i^j, \hat{q}_i^j, \theta) \label{eq:minimization} \\
    \ell_{\mathit{ans}} = \ell_{\scaleto{\mathstrut\mathit{CE}}{6pt}} (\hat{y}_{ { \mathit{start}}}, \hat{a}_{ { \mathit{start}}}) + \ell_{\scaleto{\mathstrut\mathit{CE}}{6pt}} (\hat{y}_{ {\mathit{stop}}}, \hat{a}_{ { \mathit{stop}}}) \label{eq:loss} 
\end{gather}
\noindent where $\ell_{CE}$ is cross-entropy loss.
The inference is performed on human-authored questions to predict the answer spans:
\begin{equation}
    [y_{start}, y_{stop}] = g(c, q). \label{eq:inference} 
\end{equation}

Next, we describe the variants of test-time reading comprehension.
\paragraph{Single-Context Test-Time RC.}
This is the standard formulation of test-time learning in this paper, with Equation~\ref{eq:minimization} optimizing over $\theta$, i.e.\ for each context $c_i$, the feature extractor $\theta_f$ is re-initialized with pre-trained BERT, and the answering head $\theta_h$ is randomly initialized.

\paragraph{$K$-neighbor Test-Time RC.}
In this version, $K$ contexts similar to the test-context $c_i$ are grouped together, and Equation~\ref{eq:minimization} is optimized over each set of similar contexts as opposed to single contexts in the standard setting. We index contexts in a Lucene-based information retrieval system~\cite{gormley2015elasticsearch} and retrieve top-K similar contexts given $c_i$, which we call Context Expansion with IR described in Section~\ref{selfsup_qa_gen}. 

\paragraph{Curriculum Test-Time RC.}
In the curriculum learning version, questions are ordered in increasing order of complexity. We generate different types of questions, such as, semantic role labelling, cloze-completion, template-based and dependency tree-based translation of cloze questions to natural questions. This provides an ordering of complexity and we study the effect of test-time training with such an increasing complexity. 

\paragraph{Online Test-Time RC.}
In the online test-time learning (TTLO), test samples are considered to be encountered in sequence.
As such, answering head parameters $\theta_h$ are updated sequentially without being randomly re-initialized like in the standard single-context setting.
For each new test context $c_i$, $\theta_h$ is initiliazed with the optimal parameteres from the previous test context $c_{i-1}$ to optimize Equation~\ref{eq:minimization}.

\section{Self-Supervised QA Generation}\label{selfsup_qa_gen}
In this section, we detail our framework for procedurally generating QA pairs from a given context.
We use named-entity recognition from Spacy~\cite{spacy2}, dependency parsing from Berkeley Neural Parser~\cite{stern-etal-2017-minimal} and semantic role labeling~\cite{he-etal-2015-question} as our core methods to extract plausible answers and generate natural questions. 
As described in our task formulation, we create a set of $M$ question-answer pairs $\{(\hat{q}_i^j, \hat{a}_i^j)\}_{j=1}^M$ for the given context $c_i$.

\paragraph{Cloze Generation.}
Statements in which the answer is replaced with a mask or blank token are called cloze questions. 
We follow the steps provided in~\citet{lewis-etal-2019-unsupervised} in which answers are replaced with a special token depending on the answer category.
For example, in a sentence,
\begin{quoting}
    \small
    \textit{``They were descended from Norse raiders and pirates from Denmark''}
\end{quoting}
the answer \textit{Denmark} is replaced by \textsc{[Location]}, resulting a cloze question:
\begin{quoting}
    \small
    \textit{``They were descended from  Norse raiders and pirates from \textsc{[Location]}''}.
\end{quoting}

\paragraph{Cloze Translation}
is utilized to rephrase cloze questions into more natural questions by using rule-based methods from~\citet{lewis-etal-2019-unsupervised}.

\paragraph{Template-based Question Generation}
utilizes simple template-based rules to generate questions.
Given a context of format:
\begin{quoting}
    \small
    \textsc{[Fragment A]}\textsc{[Answer]}\textsc{[Fragment B]}
\end{quoting}
a template of the format ``Wh+\textsc{B}+\textsc{A}+?''
replaces the answer with a Wh-word (e.g., who,what,where) as described in \citet{fabbri-etal-2020-template}.

\paragraph{Dependency Parsing-based Question Generation.}
In this method, we use dependency reconstruction to translate clozes to natural questions as described in~\citet{li-etal-2020-harvesting}, according to the following steps:

\noindent\begin{boxedminipage}{\linewidth}
    \begin{enumerate}[nosep,noitemsep,leftmargin=*]
        \item Right child nodes of the answer are retained and left children are pruned.
        \item For each node of the parse tree, if the  child node's subtree contains the answer, the child node is moved to the first child node.
        \item An in-order traversal is performed on the reconstructed tree. A rule-based mapping is applied to replace the special mask token of the cloze with an appropriate ``Wh-word''.
    \end{enumerate}
\end{boxedminipage}

\paragraph{QA-Semantic Role Labeling (QA-SRL)} was proposed by~\citet{he-etal-2015-question} as a method to annotate NLP data, by using QA pairs to specify textual arguments and their roles.
As seen in Figure~\ref{fig:overview}, for the context sentences: 
\begin{quoting}
    \small
    \textit{``They were descended from Norse raiders and pirates from Denmark.''}, \\
    \textit{``The distinct cultural and ethnic identity of the Normans emerged initially in the first half of the 10th century and it continued to evolve.''}
\end{quoting}
the following QA pairs were generated, 
\begin{quoting}
    \small
    \textit{(``What was someone descended from?'', ``Norse'')}, \\
    \textit{(What evolved?, distinct cultural and ethnic diversty)}
\end{quoting}
We can observe the questions are short and use generic descriptors and pronouns such as \textit{``something''} and \textit{``someone''} instead of specific references calling for the model to have greater semantic understanding of the given context.

\paragraph{Context Expansion using IR} is used in the $K$-neighbor version of TTL.
For Context Expansion, we index all paragraphs present in a Wikipedia dump in \href{https://www.elastic.co/}{ ElasticSearch}.
During test-time learning, we preprocess the context $c_i$ by removing the most frequent stop-words, and use it as a seed query to search and retrieve top-K similar contexts. 
This provides us with related paragraphs that describe similar topics, and consequently more diverse and slightly larger number of QA pairs to train compared to only $c_i$.
We then generate QA pairs using the above described methods.
We study the effect of varying the number of most similar contexts ($K$) on the downstream QA performance.

\section{Experiments}
\paragraph{Datasets.} We evaluate our learning framework on two well-known reading comprehension datasets: SQuAD 1.1~\cite{rajpurkar-etal-2016-squad} and NewsQA~\cite{trischler-etal-2017-newsqa}. 

\paragraph{QA Model.} We focus on training two transformer-encoder based models, BERT-Large~\cite{devlin-etal-2019-bert} trained with whole-word masking and DistilBERT~\cite{sanh2019distilbert}.
BERT-Large is used by current state-of-the-art methods on unsupervised extractive QA tasks and has $345$ million trainable parameters.
On the other hand, DistilBERT is a knowledge-distilled transformer-encoder based model and only has $66$ million parameters ($\sim5\times$ smaller than BERT-Large), allowing us to study the efficacy of TTL with respect to model-size.

\paragraph{Metrics.}
We use the standard metrics for extractive QA -- \textit{macro Exact Match}, where the predicted answer span is directly matched with the ground-truth, and \textit{macro F1}, which measures the overlap between the predicted and the ground-truth spans. 
For comparisons with existing unsupervised methods, since TTL operates directly on test instances, we report validation set performance only for SQuAD 1.1, as the test set is hidden. 
\paragraph{Training Setup.}
For all test-time learning variants, we limit the maximum number of questions generated per context to $4000$ and the maximum number of training steps to $1500$.
The number of training steps is linearly dependent on the selected batch size $\in [16, 64]$.
For our $K$-neighbor TTL setup that uses Context Expansion, we limit the number of retrieved contexts to $500$.
In Curriculum Test-Time RC, we ensure that all variants have an equal number ($1000$) of generated QA-pairs per-context. 
We evaluate multiple learning rates within the range 1e-5 to 5e-5. 
We use the Adam~\cite{kingma2014adam} optimizer and truncate the paragraphs to a maximum sequence length of $384$. 
The number $384$ was chosen by evaluating the $99^{th}$ percentile of the combined length of question and the contexts, to reduce training overhead and GPU memory size.
Long documents are split into multiple windows with a stride of $128$. 
All experiments were conducted on two Nvidia RTX-8000 GPUs. We use ten percent of the training data to perform three hyper-parameter trials for each variant.
We train models with three random seeds, and report the mean F1 and EM scores.

\paragraph{Baselines.}
As we generate our own data using QA-SRL, we use the following strong baselines. 
First, we train BERT-Large with generated data from previous methods described in Section~\ref{selfsup_qa_gen} and our method (which contains additional QA-SRL samples).
Second, we replicate the baselines using the low parameter-count model DistilBERT ($66$ million vs $345$ million for BERT-Large).
Third, for a fair comparison to Single-Context and $K$-neighbor test-time learning where we train models for each context independently, we propose a baseline where we train on all the test contexts together,  referred to as ``All test contexts''.
We also evaluate all TTL variants on two initializations of feature-extractor parameters -- 
\begin{enumerate}[nosep,noitemsep]
    \item ``default'' initialization of BERT-Large, i.e. $\theta_f$ pre-trained on masked language modeling and next-sentence prediction tasks, and $\theta_h$ randomly initialized for each context and trained from scratch, or
    \item $\theta_f$ and $\theta_h$ further pre-trained on $100K$ synthetic QA pairs generated procedurally using our methods described in Section~\ref{selfsup_qa_gen}   with contexts taken from the Wikipedia corpus. 
\end{enumerate}

\section{Results and Discussion}

\begin{table}[t]
\centering
\small
\resizebox{\linewidth}{!} {
\begin{tabular}{@{}lcccc@{}}
\toprule
& \multicolumn{2}{c}{SQuAD 1.1} & \multicolumn{2}{c}{NewsQA} \\
Models & Dev & Test & Dev & Test \\ \midrule
DCR~\shortcite{yu2016end} & 62.5 / 71.2 & 62.5 / 71.0 & - / - & - / - \\
mLSTM~\shortcite{krause2016multiplicative} & 64.1 / 73.9 & 64.7 /  73.7 & 34.4 / 49.6$^*$ & 34.9 / 50.0$^*$ \\
FastQAExt~\shortcite{weissenborn-etal-2017-making} & 70.3 / 78.5 & 70.8 / 78.9 & 43.7 / 56.1 & 42.8 / 56.1 \\
R-NET~\shortcite{wang2017r} & 71.1 / 79.5  & 71.3 / 79.7  & - / - & - / - \\
BERT-Large~\shortcite{devlin-etal-2019-bert} & 84.2 / 91.1 & 85.1 / 91.8 & - / - & - / - \\
SpanBERT~\shortcite{joshi-etal-2020-spanbert} & - / - & 88.8 / 94.6 &  - / - & - / 73.6 \\
DistilBERT~\shortcite{sanh2019distilbert} & 77.7 / 85.8 & - / - & 57.2 / 64.8 & 56.1 / 63.5  \\ \bottomrule
\end{tabular}
}
\caption{Results (EM / F1) from supervised methods on SQuAD 1.1 and NewsQA.}
\label{tab:compare_res_sup}
\end{table} 
\begin{table}[t]
    \centering
    \small
    \resizebox{\linewidth}{!} {
    \begin{tabular}{@{}lcccc@{}}
        \toprule
         & \multicolumn{2}{c}{SQuAD 1.1} & \multicolumn{2}{c}{NewsQA} \\
        Models & Dev & Test & Dev & Test \\ 
        \toprule
        \textit{BERT-Large} \\
        \enskip+ \citeauthor{dhingra-etal-2018-simple}$^{\dagger}$  & 28.4 / 35.8  & - / -  & 18.6 / 27.6 & 18.6 / 27.2 \\
        \enskip+ \citeauthor{lewis-etal-2019-unsupervised}$^{\ddagger}$  & 45.4 / 55.6 & 44.2 / 54.7 & 19.6 / 28.5 & 17.9 / 27.0  \\
        \rowcolor{LightGray}\enskip+ \citeauthor{li-etal-2020-harvesting}  & 62.5 / 72.6 & 61.1 / 71.4 & 33.6 / 46.3 &  32.1 / 45.1 \\
        \enskip+ \citeauthor{fabbri-etal-2020-template}  & 46.1 / 56.8 & - / - & 21.2 / 29.4 & - / -\\
        \enskip + our data & 49.4 / 59.1 & - / - & 28.2 / 37.6 & 27.3 / 36.4 \\ 
        \midrule 
        \textit{DistilBERT} \\
            \enskip + \citeauthor{lewis-etal-2019-unsupervised} data & 23.4 / 29.5 & - / -  & 14.1 / 21.6 & 14.7 / 20.6  \\
            \rowcolor{VeryLightGray}\enskip + \citeauthor{li-etal-2020-harvesting} data & 42.6 / 48.3 & - / - & 25.4 / 36.2 & 27.1 / 35.4 \\
            \enskip + \citeauthor{fabbri-etal-2020-template} data & 37.5 / 45.6 & - / - & 16.3 / 22.3 & 16.1 / 22.9 \\
            \enskip + our data & 38.9 / 46.8 & - / - & 23.2 / 31.9 & 22.4 / 31.1 \\
        \midrule
        \textit{BERT-Large} TTL$^\bigstar$ & \textbf{69.8} / \textbf{80.4 }& - / - & \textbf{38.9} / \textbf{53.2} & \textbf{38.2} /\textbf{ 52.6} \\
        \textit{DistilBERT} TTL$^\bigstar$ & \textbf{58.1} / \textbf{68.9} & - / - & \textbf{32.6} / \textbf{46.4} & \textbf{30.5} / \textbf{45.2} \\
        \bottomrule
    \end{tabular}
    }
    \caption{
    Comparison with previous unsupervised methods on SQuAD 1.1\ and NewsQA. 
    $^\bigstar$We show the best TTL model here, and results from all TTL variants in Table~\ref{tab:compare_ttl}. Metrics are EM / F1. Previous SOTA for both models are shaded in gray.
    $^*$results from~\citet{trischler-etal-2017-newsqa};
    $^\dagger$
    \citet{lewis-etal-2019-unsupervised}; 
    $^\ddagger$
    \citet{li-etal-2020-harvesting}.
    }
    \label{tab:compare_res_uqa}
\end{table}
\begin{table}[t]
    \centering
    \small
    \resizebox{\linewidth}{!} {
    \begin{tabular}{@{}lcc c cc@{}}
        \toprule
        & \multicolumn{2}{c}{Default init.\ $\theta_f$} & \hphantom{.} & \multicolumn{2}{c}{Pre-trained init.\ $\theta_f$} \\
        \cmidrule{2-3} \cmidrule{5-6}
        TTL Models & SQuAD 1.1 & NewsQA && SQuAD 1.1 & NewsQA \\
        \toprule
        {\em BERT-Large} \\
        \enskip Single-Context          & 54.9 & 34.9 & & 59.8 & 37.5 \\
        \enskip Single-Context Online   & 56.1 & 36.3 & & 61.8 & 39.1 \\
        \enskip $K$-neighbor            & \cellcolor{LightCyan}{66.2} & 41.6 & & \cellcolor{LightCyan}{78.3} & \cellcolor{LightRed}{50.7} \\
        \enskip $K$-neighbor Online     & \cellcolor{LightCyan}{\textbf{68.7}} & \cellcolor{LightRed}{46.3} & & \cellcolor{LightCyan}{\textbf{80.4}} & \cellcolor{LightRed}{\textbf{53.2}}\\
        \enskip Curriculum &       \cellcolor{LightCyan}{68.3} & \cellcolor{LightRed}{\textbf{46.7}} & & \cellcolor{LightCyan}{79.7} & \cellcolor{LightRed}{52.8} \\
        \enskip All test contexts& \cellcolor{LightCyan}{64.7} & 39.8 & & \cellcolor{LightCyan}{68.2} & 43.5\\
        \midrule 
        {\em DistilBERT} & && & &\\
        \enskip Single-Context &        37.2 & 23.2 & & 49.4 & 34.6 \\
        \enskip Single-Context Online & 38.5 & 25.3 & & 55.6 & 39.8  \\
        \enskip $K$-neighbor &        42.4 & 27.8 & & \cellcolor{LightCyan}{64.3} & 43.5 \\
        \enskip $K$-neighbor Online & \textbf{49.7} & \textbf{29.1}  & & \cellcolor{LightCyan}{\textbf{68.9}} & \cellcolor{LightRed}{\textbf{46.4}} \\
        \enskip Curriculum &      49.3 & 28.7  & & \cellcolor{LightCyan}{68.7} & \cellcolor{LightRed}{45.8}  \\
        \enskip All test contexts & 42.4 &  28.2 & &  47.4 &  38.7\\
        \bottomrule
    \end{tabular}
    }
    \caption{
    Comparison of Dev-set F1 scores for TTL variants, when $\theta_f$ are trained from default initialization for each test instance, or pre-trained on our generated data. 
    Scores surpassing previous best, are shaded in {cyan} for SQuAD and {red} for NewsQA.}
    \label{tab:compare_ttl}
\end{table}

\subsection{Unsupervised Question Answering}
We compare our results with current state-of-the-art supervised methods (Table~\ref{tab:compare_res_sup}) and unsupervised methods (Table~\ref{tab:compare_res_uqa}) on SQuAD 1.1 and NewsQA.
The previous best unsupervised method with both BERT-Large and DistilBERT is~\citet{li-etal-2020-harvesting}.
Our best TTL method is the Online version (TTLO), with a pre-training phase and a randomly-shuffled ordering of QA pairs with an average of $3000$ QA pairs per context, trained with only 100 steps.
With this setup, we are able to improve the state-of-the-art for the SQuAD benchmark with BERT-Large by $7.8\%$ exact-match accuracy and $7.3\%$ F1 score.
With DistilBERT, the best TTL method shows an improvement of $15.5\%$ EM and $20.6\%$ F1 over DistilBERT-based baseline, as shown in Table 2.
In NewsQA, TTL improves BERT-Large performance by $5.3\%$ EM and $6.9\%$ F1 score, and with DistilBERT shows an improvement of $7.2\%$ EM and  $7.2\%$ F1 score.

Training BERT-Large and DistilBERT with ``our data'' i.e.\ with a combined synthetic corpus created via all four QA-pair generation methods, marginally improves the F1 score.
This shows that our QA generation methods lead to an improvement over existing unsupervised QA generation methods as shown in Table~\ref{tab:compare_res_uqa}.
However, the TTL framework leads to even larger gains (${\sim}20\%$ for SQuAD and ${\sim}10\%$ for NewsQA), indicating the benefits of test-time learning.
This result also points to the limits of training with a large number of contexts compared to training on individual contexts.
This limitation is especially profound in lower parameter models, such as DistilBERT.
In Reading Comprehension, since the answer comes from the context, ``understanding'' the context is much more relevant. It has a higher inductive bias than learning to comprehend a significantly large number of contexts during training. 

For instance, there are multiple contexts about Normans in the SQuAD dataset, one of which is shown in Figure~\ref{fig:overview}.
But each context may have different historical persons referred to as the leaders or rulers of the Normans.
Answers to questions such as \textit{``Who was the leader of the Normans''} are better learned for each context separately than from all contexts.
Pre-training on several contexts is indeed beneficial to obtain better parameter initializations, as observed in Table \ref{tab:compare_res_uqa}, which can be further independently finetuned for each context during TTL.

\subsection{Few-Shot Question Answering}
\begin{figure}[t]
    \centering
    \includegraphics[width=0.85\linewidth]{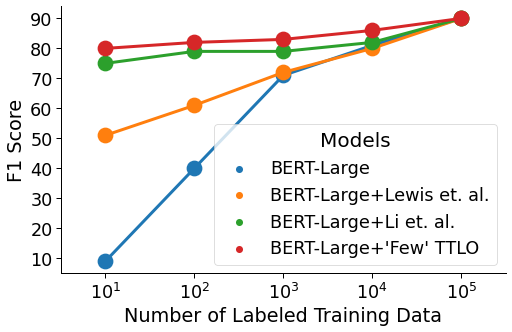}
    \caption{Comparison of F1 scores of TTL models when trained with an increasing number of labeled training samples on SQuAD. TTLO--Online TTL.}
    \label{fig:fewshot}
\end{figure}
We evaluate our best method under the few-shot setting, i.e.\ when models are trained with a limited number of human-authored QA pairs from the training datasets.
Figure~\ref{fig:fewshot} shows a comparison with an increasing number of labeled training samples for SQuAD. 
TTL-Online is consistently better than existing methods and achieves $81.6\%$ F1 score with just $100$ labeled samples.
This indicates that this learning framework can reduce the number of in-domain human-authored samples required for training.
TTL-Online is also consistently better than~\cite{li-etal-2020-harvesting} which the previous best unsupervised method for SQuAD.
All methods (which use BERT-Large as backbone) converge to similar performance, with an increasing number of additional human-authored samples.
This indicates the saturation of the inductive bias that can be incorporated into the architecture using current human-authored annotations.  

\subsection{Analysis}

\begin{table}[t]
    \centering
    \small
    \resizebox{\linewidth}{!} {
    \begin{tabular}{@{}lcc c cc@{}}
        \toprule
        Curriculum Order & \multicolumn{2}{c}{Default init.\ $\theta_f$} & \hphantom{.} & \multicolumn{2}{c}{Pre-trained $\theta_f$} \\ 
        \cmidrule{2-3} \cmidrule{5-6}
        (Left to Right) & SQuAD & NewsQA && SQuAD & NewsQA \\ \midrule
        \textit{BERT-Large}  \\
        \enskip Random  Shuffled & \underline{68.7} & 46.3 && \underline{80.4} & \underline{53.2} \\
        \enskip QA-SRL \textgreater \ T \textgreater \ DP & 68.3 & \underline{46.7} && 79.7 & 52.8 \\
        \enskip T \textgreater \ QA-SRL \textgreater \ DP & 67.6 & 45.4 && 77.6 & 50.0 \\
        \enskip T \textgreater \ DP \textgreater \ QA-SRL & 65.8 & 44.3 && 75.3 & 47.2 \\ \midrule
        \textit{DistilBERT} &  &  &  &  \\
        \enskip Random  Shuffled & \underline{49.7} & \underline{29.1} && \underline{68.9} & \underline{46.4} \\
        \enskip QA-SRL \textgreater \ T \textgreater \ DP & 49.3 & 28.7 && 68.7 & 45.8 \\
        \enskip T \textgreater \ QA-SRL \textgreater \ DP & 48.8 & 28.1 && 67.2 & 43.9 \\
        \enskip T \textgreater \ DP \textgreater  \ QA-SRL & 47.1 & 26.5 && 65.3 & 39.2 \\ \bottomrule
    \end{tabular}
    }
    \caption{
    Dev-set F1 scores for $K$-neighbor Online test-time learning, for different Curriculum Learning orderings of
    QA-SRL~\cite{he-etal-2015-question}, T (template-based methods), DP (dependency parsing).}
    \label{tab:curriculum}
\end{table}
We study the different variants of test-time learning and effects of hyperparameters, such as the number of training steps and the number of contexts, on the validation split for both datasets.

\paragraph{Single-Context vs $K$-neighbor Test-Time RC.}
In Table~\ref{tab:compare_ttl}, we compare all TTL variants.
We observe that training with additional contexts has a significant impact on F1 score, compared to training on only the given test context $c_i$.
This may be simply explained as more synthetic training samples from similar contexts leading to a better generalization to human-authored samples. 
Although similar work in image classification~\cite{sun2020test} and super-resolution~\cite{shocher2018zero} show a substantial performance improvement in a single sample learning, we observe that context expansion is beneficial for reading comprehension. 

\begin{figure}[t]
    \centering
    \includegraphics[width=\linewidth]{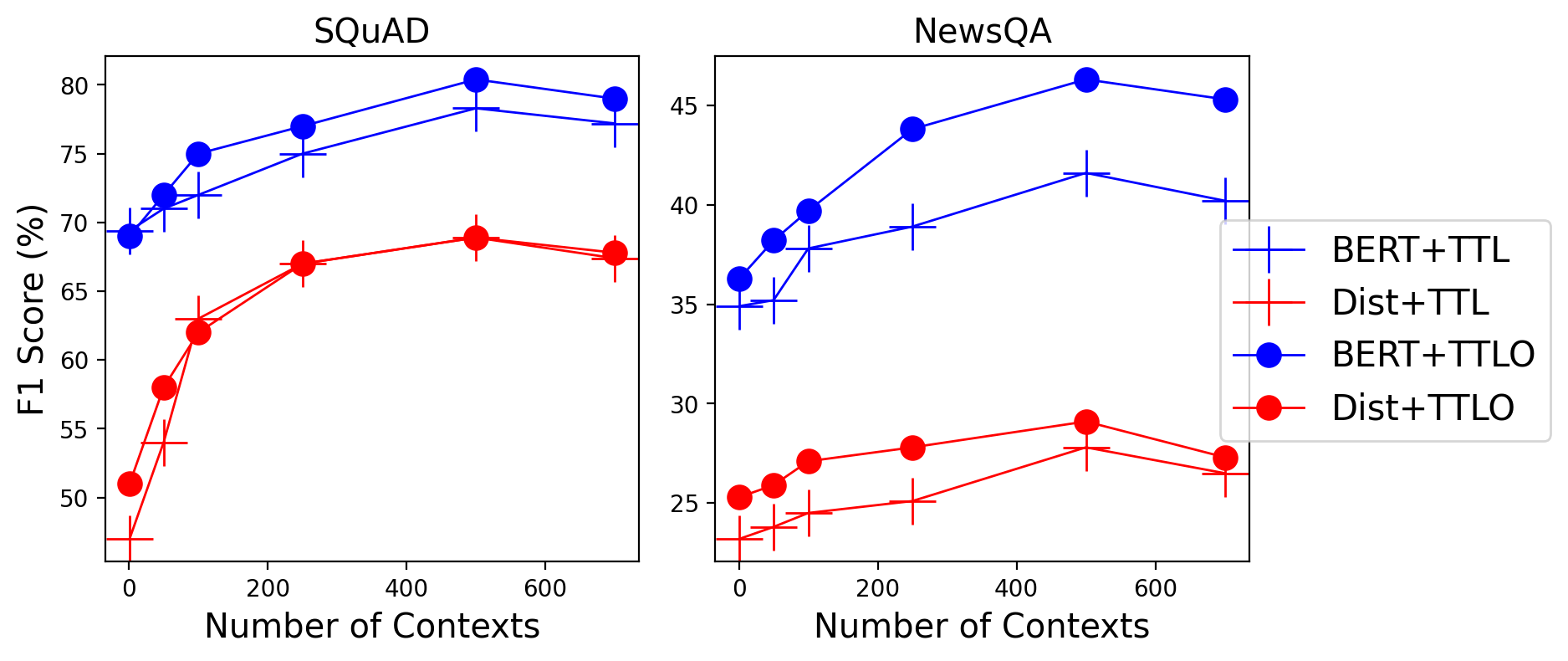}
    \caption{Comparison of F1 scores of TTL models when trained with an increasing number of contexts, on both SQuAD and NewsQA.}
    \label{fig:contexts}
\end{figure}
In Figure~\ref{fig:contexts}, we vary the number of retrieved neighbors contexts, $K$, and observe that F1 scores continue to increase till a limit ($\sim 500$).
This is consistent in both BERT-Large and DistilBERT, as well as in the two datasets, SQuAD and NewsQA. 
Our hypothesis is that there exists an optimal number of QA pairs that the model benefits from, and a maximum threshold on the number of similar contexts after which, the model starts to overfit to the synthetic nature of the QA pairs.

\paragraph{Randomly initialized v/s Pre-trained $\theta_f$,$\theta_h$.}

\begin{figure}[t]
    \centering
    \includegraphics[width=\linewidth]{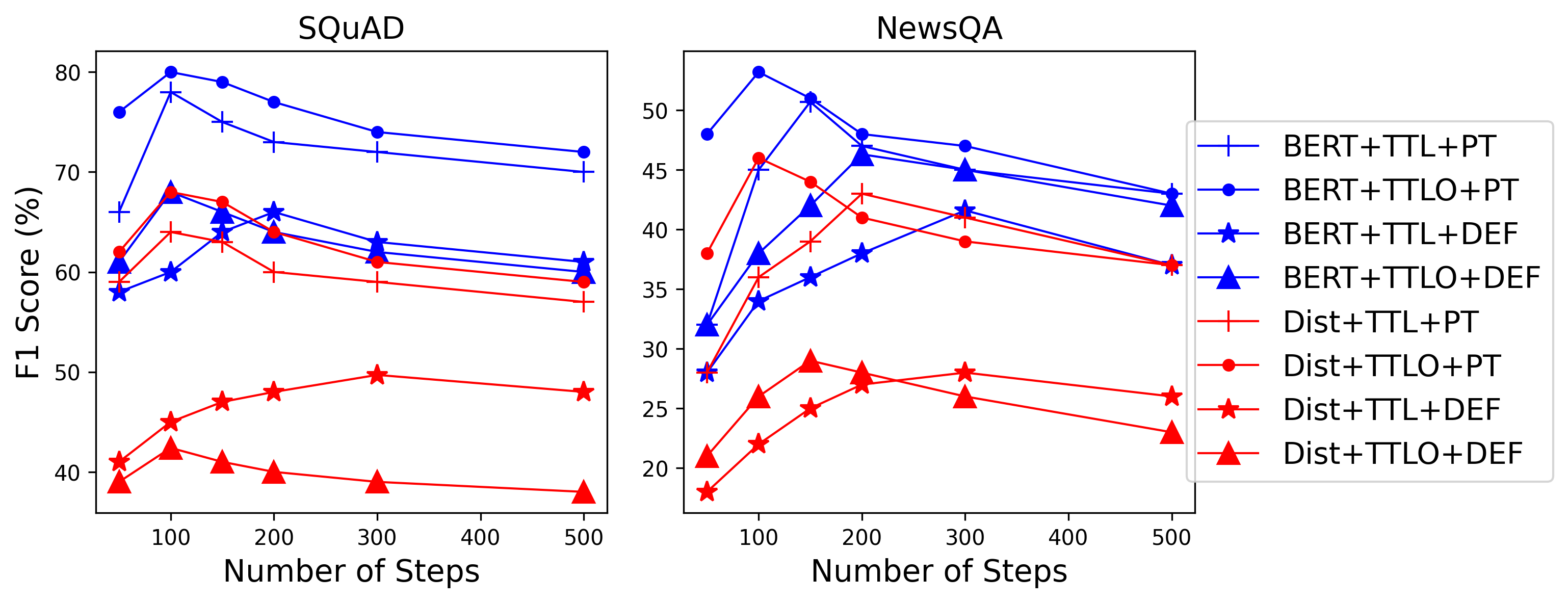}
    \caption{Effect of number of train steps on F1 scores of each TTL model on both SQuAD and NewsQA. \\PT--Pre-Trained $\theta_f,\theta_h$, DEF--Default $\theta_f,\theta_h$.}
    \label{fig:effect_learning_steps}
\end{figure}

We study the effect of re-initializing the question answering head and further pre-training using a set of procedurally generated QA-pairs on downstream test-time learning in Figure~\ref{fig:effect_learning_steps} and Table~\ref{tab:compare_ttl}.
While F1 scores achieved without pre-training are comparable to prior methods, pre-training leads to improved performance and also faster convergence, as shown in Figure~\ref{fig:effect_learning_steps}.
This can be attributed to better initial weights, which are further finetuned during the test-time learning phase.
We studied pre-training with $50k$, $100k$, and $200k$ QA pairs and observed the best performance with $100k$ samples.  

\paragraph{Curriculum Test-time learning.}
In Table \ref{tab:curriculum} we study the effect of curriculum TTL, compared to the baseline of the default random-shuffled QA pairs. 
Interestingly, using a random ordering rather than a defined curriculum begets the best performance.
Among the three curriculum ordering that we utilized, \textsc{[QA-SRL, Template-Based (T), DP (Dependency- Parsing-based)]} was effective but slightly lower than the performance with random ordering.
However, training with QA-SRL at the end has a distinctly negative effect.
We hypothesize that the model starts to overfit to the shorter vague questions from QA-SRL and ``forgets" more natural questions. Hence, it loses generalizability to the human-authored questions.

\begin{figure}[t]
    \centering
    \includegraphics[width=\linewidth]{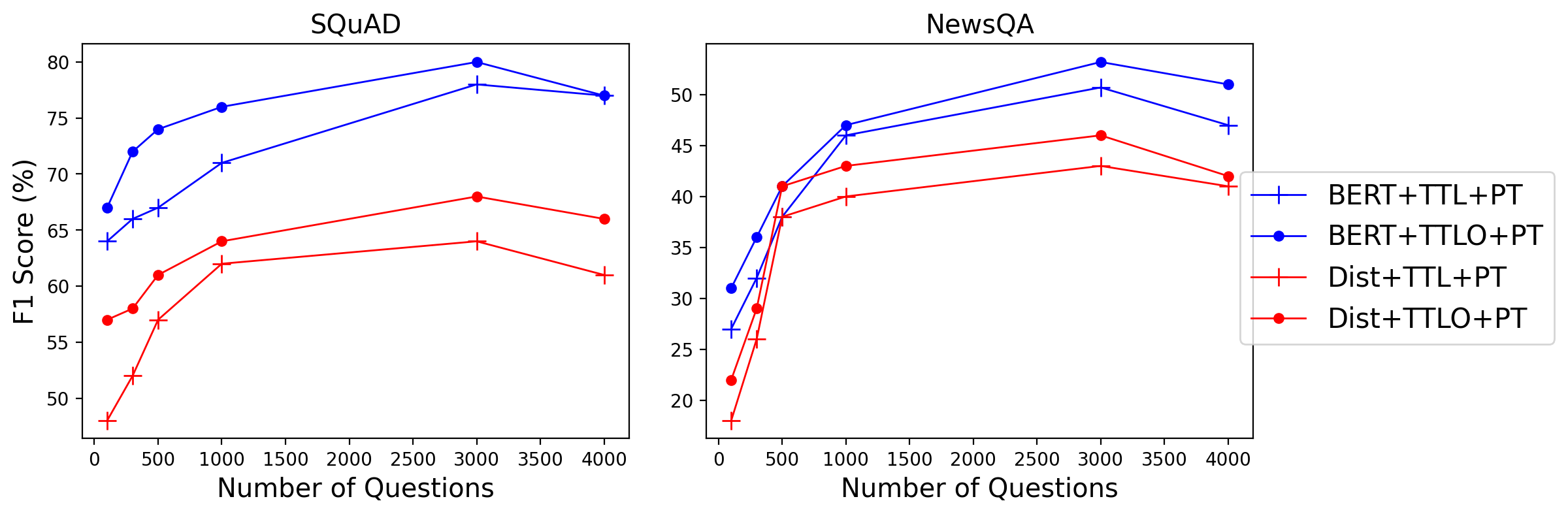}
    \caption{Effect of number of questions  on F1 scores of each TTL model on both SQuAD and NewsQA. \\PT--Pre-Trained $\theta_f$.}
    \label{fig:effect_lnum_qa_pairs}
\end{figure}

\paragraph{Online-Test-time Learning.}
In online test-time learning, the model is continuously self-supervised and evaluated on a continuous stream of contexts  and  QA-pairs.
From Table~\ref{tab:compare_ttl} and Figures~\ref{fig:contexts}, \ref{fig:effect_learning_steps} and~\ref{fig:effect_lnum_qa_pairs}, we can observe that TTL-Online consistently outperforms the single-context variant.
One key observation is that the model achieves its best performance within $100$ training steps (batch size of $48$), whereas the base version needs around $300$ to $500$ steps. This fast adaptation enables a faster inference time, compared to $\theta_h$ being trained from scratch.
We studied the effect of different random orderings of the test samples and observed the deviation as $\pm 1.6$\% in F1 scores, which indicates ordering of test samples has a minor effect.

\paragraph{Effect of Batch Size and Learning Rate.}
Batch-size and learning rate have strong effects on online test-time learning.
We observe that resuming with the learning rate of the last epoch of the pre-training with synthetic QA pairs achieves the best F1 scores.
We do not use any weight decay.
A persistent optimizer state  between contexts is critical.
Similarly, we hypothesize that the batch-layer normalization statistics pre-computed in transformer encoder layers get updated in further pre-training with QA pairs, leading to a better estimation during TTL.
For the base variant of TTL, a higher, fixed learning rate of 3e-5 with a batch size of 32-48 achieves the best F1 scores.

\paragraph{Effect of number of Training steps and QA pairs} is studied in Figures~\ref{fig:effect_learning_steps} and~\ref{fig:effect_lnum_qa_pairs}.
To limit inference time per test context, we observe TTL variants initialized with pre-trained $\theta$ achieve the top performance within $150$ training steps, whereas those trained with default initialization need $200{-}300$ steps. In Figure \ref{fig:effect_lnum_qa_pairs}, we can observe the variants achieve their best F1 scores around $3k$ QA pairs. This appears consistent with $100$ train steps with a batch size of $24{-}32$. 
Surprisingly, DistilBERT with pre-trained $\theta$ performs equally well compared to BERT-Large with no pre-training on synthetic question-answer pairs.

\paragraph{Effect of TTL on inference time.}
TTL and its variants all increase the inference time as compared to traditional inference.
For the best variant of TTL-Online with BERT-Large, we train for $100$ steps with a batch size of $48$ samples, which leads to an inference time of ${\sim}5$ minutes per context. 
Each context contains, on average $6{-}7$ questions in SQuaD 1.1 and NewsQA.
The best variant of DistilBERT, although has a lower average inference time of $1.6$ minutes per context, by employing several engineering tricks, such as saving models on RAM instead of the disk by using \texttt{tmpfs}~\cite{snyder1990tmpfs}, and using mixed-precision training~\cite{micikevicius2018mixed}.
In comparison, non-TTL methods have inference times in the range $\sim10$K samples/sec with a GPU hardware of Nvidia V100 16GB. 
TTL inference time is limited by the current computation power of the GPUs but is potentially remediable. 
However, with an increase in CUDA cores in GPUs and RAM size, we estimate the inference time can be further improved.
Moreover, with newer efficient transformer architectures such as  Linformer~\cite{wang2020linformer}
and Big Bird~\cite{zaheer2020big}, it is possible for this inference time to be further reduced.
It will be an interesting future work to increase TTL's efficiency further while retaining its strength of generalizing to evolving distributions. 

\begin{table*}[t]
    \centering
    \small
    \resizebox{\linewidth}{!} {
    \begin{tabular}{@{}@{\extracolsep{\fill}}p{90mm}>{\raggedright}ll@{}}
        \toprule
        \textbf{Question} & \textbf{Predicted} & \textbf{GT} \\
        \midrule
        What can block a legislation? & parliament & majority in parliament \\
        \midrule
        Which TFEU article defines the ordinary legislative procedure  that applies for majority of EU acts? & 294 & TFEU article 294 \\
        \midrule
        Who was killed in Dafur ? & Red Cross employee & Red Cross employee dead \\
        \midrule 
        Who does the African National Congress say should calm down ? & Archbishop Desmond Tutu & Tutu \\
        \bottomrule
    \end{tabular}
    }
    \caption{Error Analysis: Illustration of alternate plausible answers predicted by our models, but regarded as wrong predictions for SQuAD and NewsQA.}
    \label{tab:errors}
\end{table*}

\paragraph{Error Analysis.}
We analyzed $100$ wrongly answered samples from SQuAD validation split and observed the model is biased towards answering named-entities.
This is not unexpected as most of our QA-pair generation methods are focused on named-entity answers.
For example, for the question \textit{``Is it  easier or harder to change EU law than stay the same?''},  the TTL DistilBERT model generates \textit{``EU''}, whereas the ground-truth answer is ``harder''.
Although QA-SRL generates more diverse answers, the corresponding questions are vague and much more synthetic, leaving scope for improving QA pair generation to include a variety of question and answer types in the future.
Another source of errors is the alternate plausible answers generated by our models, shown in Table~\ref{tab:errors}.


\section{Related Work}
\paragraph{Extractive QA.}
The goal for extractive question answering (EQA) is to predict a span of text in a context document as the answer to a question.
Various benchmarks have been established to evaluate the capability of EQA models on corpuses from different domains such as Wikipedia-based question answering in  SQuAD~\cite{rajpurkar-etal-2016-squad}, Natural Questions dataset~\cite{kwiatkowski-etal-2019-natural}, as well as questions requiring complex reasoning to extract answers in HotPotQA~\cite{yang-etal-2018-hotpotqa};
questions about news' articles in NewsQA~\cite{trischler-etal-2017-newsqa};
and about trivia-facts in TriviaQA~\cite{joshi-etal-2017-triviaqa}.

\paragraph{Unsupervised QA.}
For many of the aforementioned extractive QA benchmarks, ``human-like'' performance has been reached via supervised methods.
Unfortunately, these methods do not transfer well to new domains, and the collection of training data in new domains and new languages may not always be feasible.
To address this, unsupervised EQA has been proposed as a challenge~\cite{lewis-etal-2019-unsupervised}, in which aligned {\it(context, question, answer)} triplets are not available.
Self-supervised data-synthesis methods~\citep{lewis-etal-2019-unsupervised,banerjee-baral-2020-self,rennie-etal-2020-unsupervised,fabbri-etal-2020-template,li-etal-2020-harvesting, banerjee2020self} have been used for question answering by procedurally generating QA pairs and training models on these synthetic data.  

\paragraph{Self-Supervised Learning.}
The key idea in self-supervision is to design auxiliary tasks so as to  and extract semantic features from unlabeled samples, for which input-output data samples can be created from unlabeled datasets.
Self-supervision has been used to train large transformer-based language models such as BERT~\cite{devlin-etal-2019-bert} and T5~\cite{2020t5} for the auxiliary task of masked token prediction, and XLNET~\cite{yang2019xlnet} for token prediction given any combination of other tokens in the sequence.
ELECTRA~\cite{clark2019electra} instead of masking tokens, jointly trains a generator to substitute input tokens with plausible alternatives and a discriminator to predict the presence or absence of substitution.
MARGE~\cite{lewis2020pre} is trained to retrieve a set of related multi-lingual texts for a target document, and to reconstruct the target document from the retrieved documents. 
The goal of self-supervised pretext task design is to come up with tasks that are as close to the main task, to learn better representations. 
In NLP, QA format provides us such an opportunity where we can leverage NER, SRL, Cloze Completion as auxiliary tasks for complex QA.

\paragraph{Learning at test-time.}
Our work is inspired by image processing methods such as single-image super-resolution~\citep{glasner2009super,freedman2011image,shocher2018zero} that do not require access to external training datasets but instead formulate a self-supervised task for upsampling natural image patches recurring at different scales in the image.
Test-time training (TTT)~\cite{sun2020test} for image classification makes use of rotation prediction~\citet{gidaris2018unsupervised} as an auxiliary task to implicitly learn image classification at test-time and shows improved robustness.
While we can directly synthesize main-task data (QA pairs) from the context and do not require an auxiliary task, our work is closely related to TTT.

\paragraph{Domain Adaptation.} 
Pre-training for the tasks such as masked language modeling or other synthetic tasks on unlabeled corpora for a new domain has been evaluated for commonsense reasoning~\cite{mitra2019additional} and classification tasks~\cite{gururangan-etal-2020-dont}.
On the other hand, our work can be viewed as task-specific self-supervision with each new context as a new domain. 

\section{Conclusion}

In this work, we propose test-time learning (TTL) as a new framework for unsupervised extractive question answering (EQA). We present four variants of TTL with a simple but effective context expansion method. We utilize four question-answer pair generation methods for EQA and propose using QA-SRL as an additional source of QA pairs, to supplement prior methods. We show TTL enables ``understanding'' of contexts at test-time, without human-authored annotations, and significantly improves EQA, including low parameter models. 

We envision TTL as a framework that can direct work in reading comprehension to be viewed as a problem of ever-evolving datasets instead of a static corpus. 
Natural language itself undergoes continuous evolution~\cite{gentner1988verb,traugott2001regularity,hamilton-etal-2016-cultural} via changes in preference for syntactical structures; creation of new words and phrases; and changing usage frequencies and semantics for existing words.
TTL can potentially be applied to such scenarios with semantic drift or domain shift. 
Further improvements w.r.t.\ selection of similar contexts for K-neighbor TTL could be explored by leveraging hard sample selection, hard negative mining, bootstrapping, and contrastive learning, along with improved currculum strategies.

\section*{Acknowledgements}
The authors acknowledge support from the DARPA SAIL-ON program W911NF2020006, ONR award N00014-20-1-2332 and NSF grant 1816039; and thank the reviewers for their feedback. 

\section*{Ethical Considerations}
Our test-time learning method treats every new test instance as a new distribution, and does not rely on a human-authored training dataset.
We believe that this is a possible way to avoid learning spurious correlations or linguistic priors, especially when it comes to socio-cultural and historical biases that have been shown to percolate into models for various NLP tasks~\cite{hendricks2018women,kurita-etal-2019-measuring,sheng-etal-2019-woman}.
On the other hand, if the test context itself contains biased, false, or propaganda statements, our model will use those statements to extract answers.
We would not want models trained on such data to be deployed in the real world.
However, because model parameters are randomly initialized for each new context in the standard version of our framework, if contexts are fact-checked by ``reliable'' sources, then we believe our model will be relatively bias-free, as compared to pre-trained language models for which it is hard to trace \textit{why} a certain prediction was made. 
Test-time learning allows us to disentangle biases learned from single contexts, from biases learned by language models from large corpora.

\bibliography{anthology,custom}
\bibliographystyle{acl_natbib}

\appendix

\end{document}